\documentclass[letterpaper]{article} 
\usepackage{aaai23}  
\usepackage{times}  
\usepackage{helvet}  
\usepackage{courier}  
\usepackage[hyphens]{url}  
\usepackage{graphicx} 
\usepackage{subcaption}
\usepackage{natbib}  
\usepackage{caption} 
\usepackage{algorithm}
\usepackage{algorithmic}

\usepackage{newfloat}
\usepackage{listings}
\DeclareCaptionStyle{ruled}{labelfont=normalfont,labelsep=colon,strut=off} 
\floatstyle{ruled}
\newfloat{listing}{tb}{lst}{}
\floatname{listing}{Listing}
\title{Reconstructing Existing Levels through Level Inpainting}

\author{Johor Jara Gonzalez, Matthew Guzdial}
\affiliations{Computing Science Department, Alberta Machine Intelligence Institute\\
University of Alberta\\
jaragonz@ualberta.ca, guzdial@ualberta.ca
}

\usepackage{bibentry}

\begin{document}

\maketitle
\begin{abstract}
\begin{quote}
Procedural Content Generation (PCG) and Procedural Content Generation via Machine Learning (PCGML) have been used in prior work for generating levels in various games. This paper introduces Content Augmentation and focuses on the subproblem of level inpainting, which involves reconstructing and extending video game levels. Drawing inspiration from image inpainting, we adapt two techniques from this domain to address our specific use case. We present two approaches for level inpainting: an Autoencoder and a U-net. Through a comprehensive case study, we demonstrate their superior performance compared to a baseline method and discuss their relative merits. Furthermore, we provide a practical demonstration of both approaches for the level inpainting task and offer insights into potential directions for future research.
\end{quote}
\end{abstract}

\section{Introduction}

Procedural Content Generation (PCG) \cite{pcgsurvey} and Procedural Content Generation via Machine Learning (PCGML) \cite{Summerville2017}, cover techniques used to generate different types of content, such as ``levels, maps, game rules, textures, stories, items, quests, music, weapons, vehicles, characters, etc" \cite{pcgbook}. 
While PCG requires hand-authored knowledge from a developer, PCGML instead extracts this knowledge from existing content. 
One problem with PCGML is the question of where it slots into the game development process. 
Early in a game's development there won't be sufficient content to train a model. 
However, if we wait until the end of the development process then there's no point in applying PCGML, as all the content has already been produced. 
But if we consider instead not helping with the initial development of the game, but applying PCGML to extend or augment an already completed game, we might sidestep this issue.

We establish a novel PCG problem called Content Augmentation (CA), the main characteristic of CA is that it extends original content. 
CA might, for example, be used in the generation of DLC, mods, or updates in the case of a live service game. 
We'll discuss CA applications further in the future work section at the end of the paper. 
The difference between CA and other applications of PCG is that they tend to focus on generating content from scratch, based on user input or hand authored rules. 
In comparison, CA must take into account the existing work that it seeks to extend or augment. 

Many branches of PCG have some overlap with CA.
Classical PCG approaches like search or grammars could be applied to CA. 
But these approaches do not traditionally focus on creating content based on some designer's prior work \cite{pcgbook}. 
PCGML techniques have attempted to learn to emulate a particular designer, however this has thus far been focused on generating whole new pieces of content with that same style, rather than extending existing content \cite{Summerville2017}.

Some co-creative systems \cite{deterding2017mixed}, where a human works with a PCG system to produce content, do attempt to extend or modify output from a human, but typically as part of an initial design process. 
In comparison, CA focuses on taking complete, output content and extending it after it has been published or released. 
PCGML techniques have been proposed to try to generate whole new games, a problem sometimes called automated game generation \cite{snodgrass2020multi,Guzdial2018}.
This is related to CA, in terms of taking as input existing games, but we focus on augmenting or extending game content instead of generating whole new games from scratch.

Rather than attempting to solve CA as a single problem, we identify and attempt to solve an initial CA subproblem that we call level inpainting. In level inpainting we focus on reconstructing the existing structures in a game. 
Level inpainting is directly inspired by image inpainting \cite{ImageInpainting}, a computer vision problem domain focused on reconstructing missing or damaged structures in images.
Similarly, we aim to train a model to reconstruct damaged or missing structures in levels. 
While prior work has included models capable of repairing levels \cite{jain2016autoencoders}, it has focused on making changes to  the levels to make them playable. 
Instead, we focus on learning the style of a set of game levels and then filling in missing, ``masked out'' sections of other levels. 
Our future goal is to use these models to expand levels or to create new level structure that can connect disparate areas as part of a larger CA system.

As we mentioned above, for level inpainting we take inspiration from image inpainting \cite{imageinpaintingsurvey}. ``Image inpainting is a task of reconstructing missing regions in an image'' \cite{ImageInpainting}. We adapted two existing image inpainting techniques to our level inpainting task, changing them to model level structure instead of pixels.
While there has been a great deal of prior image inpainting work \cite{ImageInpainting}, it is not the case that we can just adapt them to game content naively due to a number of factors. For example, unlike images, game content must have certain characteristics to be valid (e.g. playable for levels).

In this paper, we introduce level inpainting: the process of fixing or restoring missing parts of a level. We consider this to be an initial exploration of Content Augmentation (CA). 
We validate our approach on existing game levels, by comparing the output (inpainted level) and the original level. 
As an initial exploration, we adapt two existing image inpainting approaches to level inpainting.

This paper includes the following contributions:
\begin{itemize}
\item{} We propose Content Augmentation (CA) as a PCG problem.
\item{} We propose level inpainting as a CA subproblem.
\item{} We present an approach to process an existing dataset for level inpainting.
\item{} We present two modifications of image inpainting architectures adapted to level inpainting.
\item{} We present our results in comparison to a traditional PCGML baseline.
\end{itemize}

\section{Related Work}

We imagine that classical PCG approaches like search-based and constructive PCG could be applied to Content Augmentation (CA) \cite{pcgbook}. 
However, in this paper we focus on Procedural Content Generation via Machine Learning (PCGML) \cite{Summerville2017}, as we argue that CA can serve as a natural way to include PCGML in the game development process.

\textit{Super Mario Bros.} has been one of the  most common problem domains in PCGML. From generating levels \cite{summerville2016super}, blending levels \cite{sarkar2021generating}, reinforcement learning \cite{shu2021experience}, or to studying new representations for level generation \cite{jadhav2021tile}. 
There are different factors that make \textit{Super Mario} levels useful to test different approaches. One factor is that it is a finished product, for this reason the outputs of the models are easy to compare with the original content. Another factor is the popularity of the game, Mario is well recognized and the majority of people know how it is played. 
Because of these reasons, we decided to employ Mario as the problem domain for this initial experiment into level inpainting. 
The fact that we have the original levels for comparison will allow us to evaluate our approach quantitatively. Further, we hope that Mario's well-known design style will help clarify the problem of level inpainting to readers.

Wave Function Collapse (WFC)\cite{kim2019automatic} is one particular PCGML algorithm. 
WFC focuses on extracting local patterns and then using the extracted patterns to incrementally generate an output that, in our opinion, is surprisingly consistent with the original author's style.
Researchers have previously explored the generation of level designs using WFC. Some of those works include Sandhu et. al. \cite{Sandhu2019} who add design constraints to WFC in order to use them at runtime to design levels. Similarly Kim et. al. \cite{kim2019automatic} add more constraints to the original WFC approach in order to create multi-layer levels. 
In order to use WFC some works like Cheng et. al. \cite{cheng2020automatic} use a graph-based input to generate more content. 
Due to it being based on an incremental improvement of an initially random piece of content, one might consider WFC to be a natural fit for level inpainting.
We do not draw on WFC in this work, as it is the initial exploration of level inpainting and so we instead focus on adapting image inpainting models. 
However, we hope to explore WFC for level inpainting in future work.

Recently, diffusion-based approaches have achieved great success in generating images \cite{nichol2021glide}. Diffusion works by essentially learning to reverse a function that converts structured data (such as images) into noise \cite{song2019generative}.
Recently, Siper et al. \cite{siper2022path} introduced a similar idea, but applied it to level generation.
Similar to WFC, this style of approach iteratively converts an apparently random initial state into structured output, making it a good fit for level inpainting. To our knowledge, there are no existing PCGML implementations for Content Augmentation. The closest prior work has attempted to generate new content for new games with PCGML \cite{Sarkar2018, Guzdial2018,Snodgrass2020}. 
For this work, in comparison to this prior work, we aim to train a model on only one domain. 
In addition, we are not trying to create new content from scratch, instead we are focusing on reconstructing areas of the same levels using observed similarities with the rest of the map. We use this level inpainting as a proxy for extending existing levels, which we also demonstrate below as an example.

Level repair is a related PCG problem that focuses on altering levels such that the level matches some mechanical (e.g. playable) or aesthetic requirements \cite{jain2016autoencoders,cooper2020pathfinding,recgan}.
Jain et al. demonstrated level repair by mapping some unplayable section of a game level to the closest playable level section \cite{jain2016autoencoders}.
In comparison, Cooper and Sarkar employed an agent-based approach, which attempted to pathfind across a level and could alter the level to make it playable \cite{cooper2020pathfinding}. In another approach, Zhang et al. used generative adversarial networks to “generate-then-repair” using hand-author maps to train the model, then generate new maps that may be unplayable and repair them \cite{recgan}.
These approaches were altering content, but for the purpose of level playability, not to match a style or replace an undefined section.

Image inpainting \cite{imageinpaintingsurvey} uses different image processing techniques to reconstruct damaged areas of an image, a variety of approaches have been explored including Encoder-decoders \cite{liu2019coherent}, CNNs \cite{liu2019coherent}, and GANs \cite{demir2018patch}.  
The principal characteristic of image inpainting is that in order to reconstruct a damaged image it uses nearby pixels to the corrupted part to achieve coherence with the reconstructed image. In our research we modify two image inpainting architectures, so that instead of working with pixels they work with tiles. 
The goal of level inpainting then is to fill in damaged or missing tiles in order to achieve coherence with the entire level.

\section{System Overview}

To reach our goal of level inpainting, we cannot directly use the architectures of image inpainting models because they are made to reconstruct pixels in an image. Given that we want to capture the same style of the original game, we need to adapt an image inpainting architecture to our needs. 
In order to get a functional system first we need a level inpainting dataset we can use to train our models. To get a level inpainting dataset, we make use of an existing tile-based representation \cite{Summerville2016}, but post-process the data to fit the requirements of our level inpainting problem. 
Finally, we train two models adapted from image inpainting to level inpainting: an autoencoder and a U-net.

\subsection{Dataset}

For image inpainting, the datasets consist of images. 
However in our approach, using only images of the level will be impossible, as if we use pixels we cannot load the results into a game engine. 
Alternatively, we could post-process the output of our model to convert from pixels to tiles, but this would create undue complexity. 

For our dataset we make use of the Video Game Level Corpus (VGLC) \cite{Summerville2016}. The VGLC contains a tile representation of levels that can be easily adapted for our purpose of level inpainting. 
Our goal for level inpainting is to extend an author's design with the same style. 
We wanted to use a set of levels with a clear design vision. 
With this in mind, we chose to use the tile representation from \textit{Super Mario Bros.} and \textit{Super Mario Bros. 2} (the Japanese version), given that both games were made by the same designer. We split the dataset into 26 levels for training and 6 levels for testing, selecting the longest levels across the dataset for the test set (From Super Mario: 3-1, 6-2, 8-1, and from Super Mario 2: 1-1, 5-2).
We chose to use whole levels for our test set as this will let us investigate whether our models have learned the level design style, not just replicating a particular level's geometry.
If we did not use whole levels but had the same level spread across our train and test sets, then the model would see the same patterns across both sets. 
With this mind, the selection of the longest levels gave us the most opportunities to test how our models performed on different structures.

\subsection{Data Processing}

\begin{figure}[tbh]
    \centering
    \includegraphics[scale=0.6]{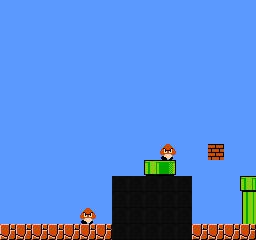}     
    \caption{Size of the mask in one example. We note that not all parts of the pipe are fully covered, which requires the model to reconstruct the missing tiles.}
    \label{fig:mask}
\end{figure}

For our data processing first we represent each level as a matrix of the same size as the level. 
The information inside of the matrix is based on a one-hot encoding representing the tile at that position. 
The one-hot encoding used is of length 13 given the tiles representation for \textit{Super Mario Bros.} in the VGLC (e.g. brick, emptyBlock, enemy,topLeftPipe, leftPipe, etc.). 
The level can thus be represented as a matrix with a height of 16, a length of the size of the level, and a depth of 13 for the one-hot encoded representation.
Then, in order to get our final training data we split each matrix into submatrices of size 16x16x13. 
With this step we get around the varying width of the levels. The representation for the sub matrices 16x16x$D$ where $D$ is the dimension of a one-hot encoding is a common representation in PCGML \cite{yang2020game}.

Once we have all the submatrices processed, we introduce a mask area for each submatrix. 
We inherit the concept of a mask from image inpainting \cite{ImageInpainting}.
In the area of the mask, we delete the information of the tiles and replace it with all zeros.
The mask has a size of 5x4 and we produced 11 masked inputs for every submatrix, by placing the 5x4 mask along each possible position at the bottom of the submatrix.
We chose this setup for our mask for several reasons. 
First, it has a reasonable size that allows it to at least partially cover common structures like pipes and stairs. Second, we place the mask along the bottom of the level to be able to cover these structures. Third, in the game \textit{Super Mario Bros.} the majority of structures and movement are located along the bottom of the level. 
In Figure \ref{fig:mask} we can demonstrate an example of the mask, and the area that the model will predict.
We note that finding useful shapes and positions for masks is an open resource problem for image inpainting \cite{ImageInpainting}, and so the same will likely be true for level inpainting.

\subsection{Model Architecture}

\begin{figure}
\subcaptionbox*{\centering  a) Convolutional autoencoder architecture }[.9\linewidth]{
    \includegraphics[scale=0.19]{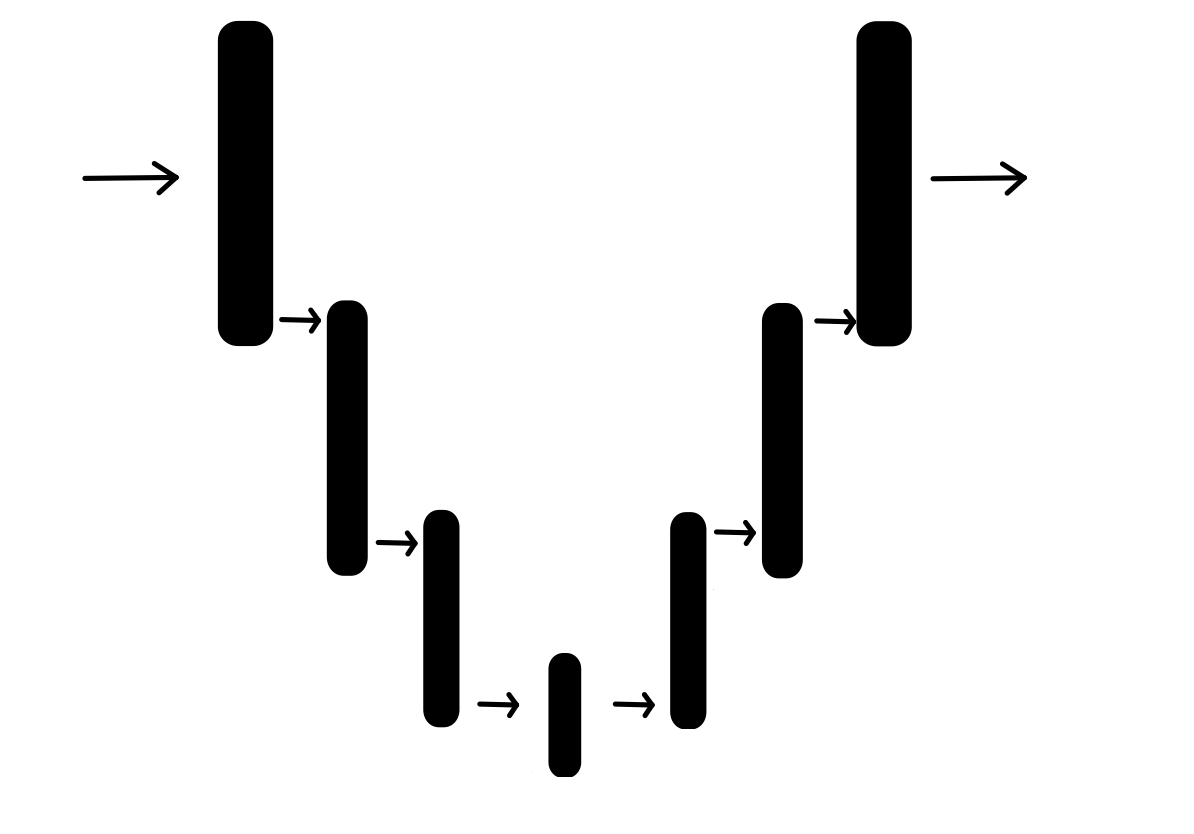}
  }
\subcaptionbox*{\centering  b) U-net architecture}[.9\linewidth]{
    \includegraphics[scale=0.1]{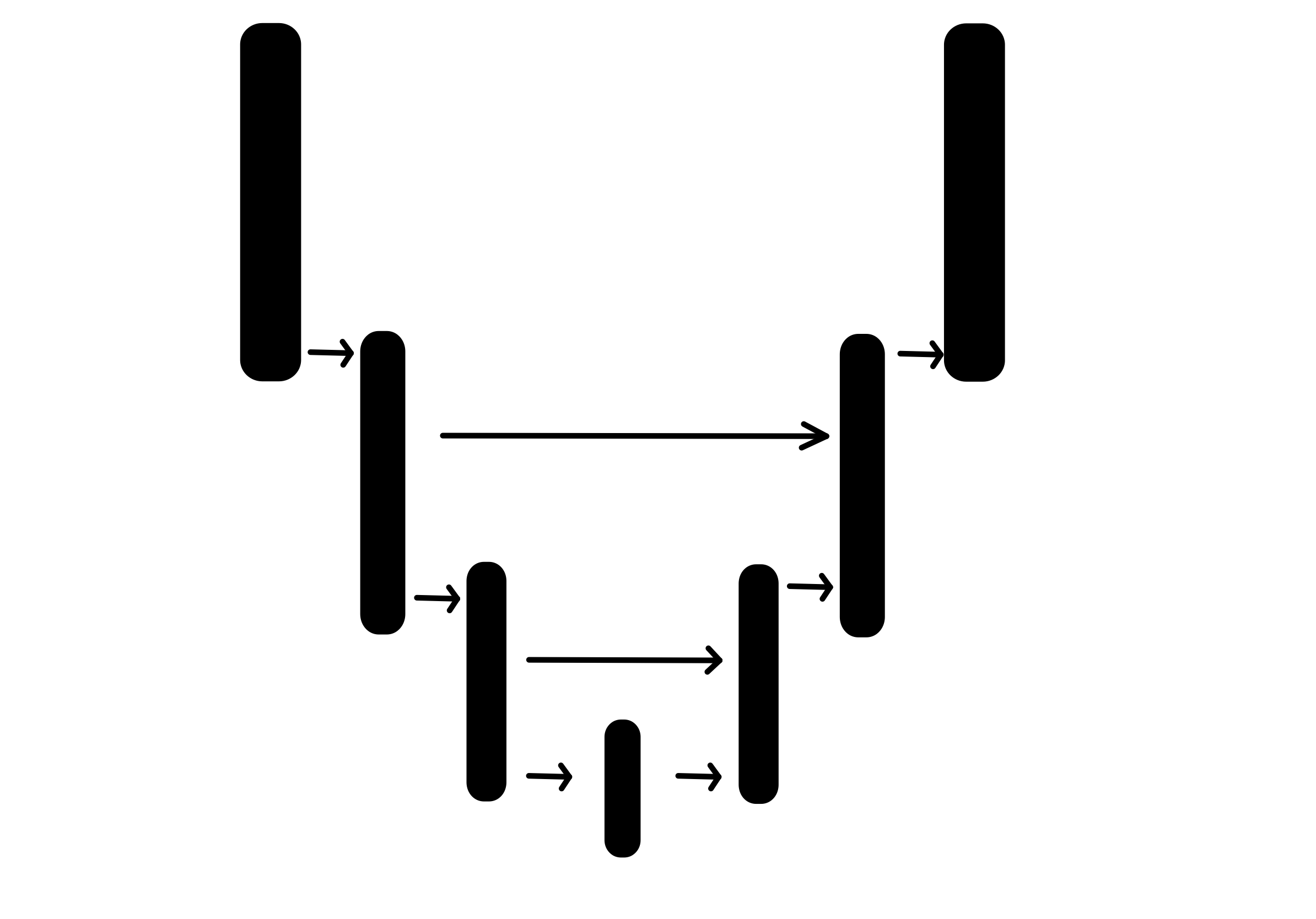}
  }

\caption{Our Two proposed models. a): the convolutional autoencoder. b): the U-net. Layers of ([16x16x13],[16x16x16],[8x8x32],[8x8x64]).}
\label{fig:example}
\end{figure}

In this section we discuss how we adapted two image inpainting architectures for level inpainting. 
The first architecture is based on a convolutional autoencoder \cite{zhang2018better}, as it is a basic architectures for image processing, including image inpainting. 
Autoencoders have been commonly applied in PCGML \cite{thakkar2019autoencoder,snodgrass2020multi,sarkar2021generating}, as they tend to be less data-hungry than other models. 
The second architecture proposed is a U-net \cite{ronneberger2015u} architecture, which was designed explicitly for image inpainting.
To have a fair comparison between our alternative models we used the same number of convolutional layers for both models. However, the U-net architecture was specifically designed to use small amounts of data, which is often the case for PCGML applications \cite{Summerville2017}.  

\begin{figure*}[tbh]
    \centering
    \includegraphics[width=5in]{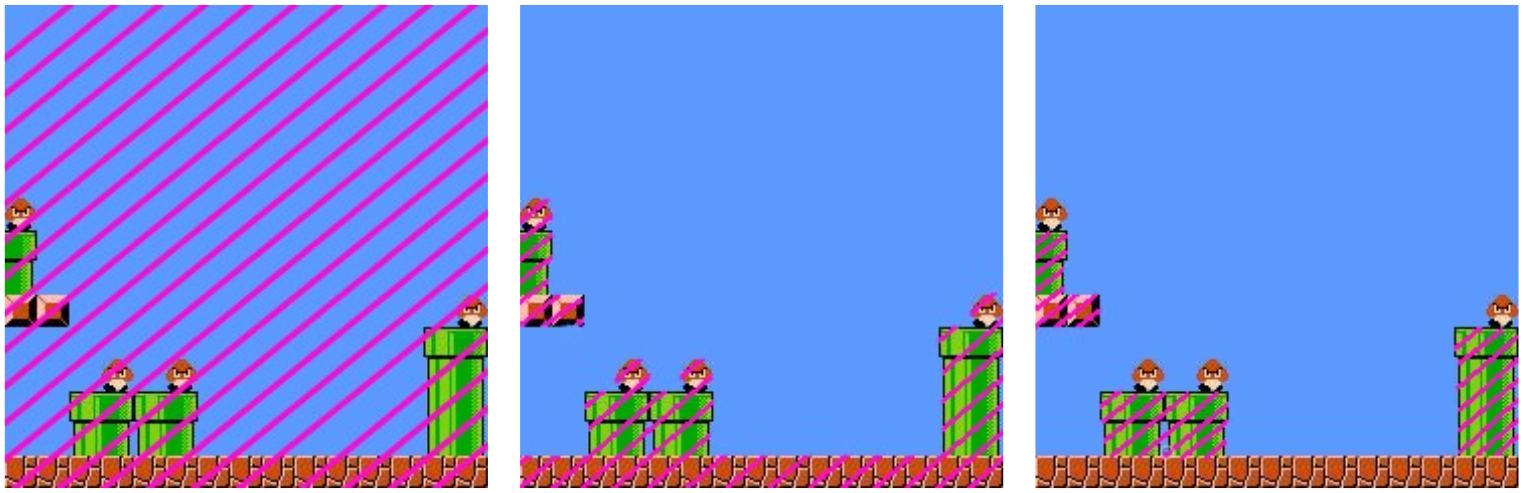}    
    \caption{A visualization of our three metrics: the Tile-by-Tile (Left), No Sky (Middle), and Structures (Right).}
    \label{fig:metrics}
\end{figure*}

Our architectures are illustrated in Figure \ref{fig:example}, both contain two paths: an encoder path with layers shown in that figure and a decoder path with the same layers but mirrored. 
Figure \ref{fig:example}a is a relatively simple convolutional autoencoder, but this is appropriate for the problem and amount of training data, as supported by prior PCGML approaches using autoencoders.
Figure \ref{fig:example}b demonstrates the U-net with the same layers, which supplements the encoding and decoding paths by propagating context to higher resolution layers.

In order to improve our architectures, we opted to use transpose convolution layers. We found they outperformed the up-conv layers typically used in image inpainting according to our experiments.
We anticipate that this is because the extra convolution made by the transpose gave us clearer results, which may be due to the differences in level and tile inpainting. 
In image inpainting, being a single pixel off won't have a large impact in comparison to being a single tile off in level inpainting.
We used binary cross entropy as a loss function for both models. 
We used the Adam optimizer with a learning rate of 0.0001, with a batch size of 10 (using as reference the quantity of submatrices generated in our dataset), and trained both models until convergence.

\begin{table*}[]
\centering
\resizebox{\textwidth}{!}{%

\begin{tabular}{|c|c|c|c|c|c|c|c|}
\hline
Metrics &
  SM1-Level 3-1 &
  SM1-Level 4-2 &
  SM1-Level 6-2 &
  SM1-Level 8-1 &
  SM2-Level 1-1 &
  SM2-Level 5-2 &
  Avg \\ \hline
AE-TbyT &
  \textbf{91.86 $\pm$ 0.42} &
  \textbf{84.07 $\pm$ 0.94} &
  \textbf{88.65 $\pm$ 0.71} &
  \textbf{92.02 $\pm$ 0.23} &
  \textbf{91.82 $\pm$ 0.25} &
  \textbf{79.49 $\pm$ 0.90} &
  \textbf{88.03} \\
UNet-TbyT &
  88.44 $\pm$ 1.12 &
  78.75 $\pm$ 1.12 &
  87.13 $\pm$ 0.67 &
  91.04 $\pm$ 0.21 &
  91.20 $\pm$ 0.30 &
  75.80 $\pm$ 0.80 &
  85.39 \\
Markov-TbyT &
  75.44 $\pm$ 5.45 &
  67.58 $\pm$ 4.58 &
  71.21 $\pm$ 2.56 &
  80.23 $\pm$ 2.56 &
  79.88 $\pm$ 2.57 &
  47.36 $\pm$ 9.40 &
  58.24 \\ \hline
AE-NoSky &
  84.75 $\pm$ 0.88 &
  \textbf{74.32 $\pm$ 2.52} &
  \textbf{77.68 $\pm$ 2.52} &
  84.24 $\pm$ 0.792 &
  \textbf{84.36 $\pm$ 0.73} &
  \textbf{63.98 $\pm$ 2.87} &
  \textbf{78.10} \\
UNet-NoSky &
  \textbf{87.20 $\pm$ 0.93} &
  66.36 $\pm$ 1.40 &
  75.97 $\pm$ 1.40 &
  \textbf{85.73 $\pm$ 0.66} &
  83.93 $\pm$ 1.00 &
  54.76 $\pm$ 1.42 &
  75.65 \\
Markov-NoSky &
  53.17 $\pm$ 3.10 &
  41.76 $\pm$ 4.77 &
  48.29 $\pm$ 2.67 &
  60.89 $\pm$ 0.93 &
  57.73 $\pm$ 10.42 &
  35.14 $\pm$ 2.86 &
  49.49 \\ \hline
AE-Struct &
  82.98 $\pm$ 2.87 &
  79.24 $\pm$ 2.61 &
  56.43 $\pm$ 2.07 &
  83.44 $\pm$ 2.65 &
  \textbf{79.60 $\pm$ 1.31} &
  \textbf{57.30 $\pm$ 6.05} &
  73.16 \\
UNet-Struct &
  \textbf{88.31 $\pm$ 0.57} &
  \textbf{80.54 $\pm$ 1.54} &
  \textbf{59.76 $\pm$ 1.30} &
  \textbf{87.70 $\pm$ 1.42} &
  76.76 $\pm$ 1.89 &
  48.19 $\pm$ 3.37 &
  \textbf{73.54} \\
Markov-Struct &
  0 &
  1.53 $\pm$ 0.10 &
  2.74 $\pm$ 0.22 &
  0 &
  0 &
  0 &
  0.71 \\ \hline
\end{tabular}%
}
\caption{This table contains the the average and standard deviation values for our three different metrics across our two approaches and baseline. All three of the metrics are based on a per-tile accuracy, but focus on increasingly particular tile types.}
\label{tab:t1}
\end{table*}
\section{Evaluation}

Since our final goal is to apply level inpainting to Content Augmentation (CA), for our evaluation we would ideally want to create augmentations to Mario levels and then see how players react to these augmentations.
But for an initial exploration of this task, we're instead focused on an approximation of this problem where we know the ground truth. 
We have our models attempt to recreate existing Mario level structure and then compare their performance to a baseline. A good result would be to perfectly reproduce the area of the mask. 
In our evaluation we only focus on the interior of the mask, because we are not concerned with representational ability generally. 
We only care about a level inapinting model's ability to do the required level inpainting task.

We use three different metrics in our evaluation:
\begin{enumerate}
    \item \textbf{Tile-by-Tile (TbyT)}: First, we compare tile-by-tile accuracy. This is a standard accuracy measure used in prior PCGML work \cite{Summerville2017}. 
    Essentially, we are measuring how close the output of the approach is to the original considering all tiles.
    \item \textbf{No Sky (NoSky)}: The second metric is again a tile-by-tile comparison but we exclude the sky tiles. 
    Since the majority of the samples contains sky tiles a simple strategy would be to always predict sky tiles. So here we measure the non-sky accuracy. 
    \item \textbf{Structures (Struct.)}: The third metric, is based on the tile accuracy for the pipe and stair tiles only. This is due to the fact that these structures are some of the more complicated parts of \textit{Super Mario Bros.} levels to reconstruct, and because they are used to test the mechanics of the game (e.g. jumping, running, etc.). Further, checking for ``broken'' pipes is a common PCGML metric \cite{summerville2016super,snodgrass2014experiments}, which this approximates.
\end{enumerate}

\subsection{Baseline}

In order to better understand the results for the convolutional autoencoder and the U-net, we implement a Markov chain baseline.
We chose a Markov baseline because it is one of the most common and oldest PCGML approaches for \textit{Super Mario Bros.} levels \cite{snodgrass2016controllable}. 
In addition, given that a Markov chain was the first PCGML approach for level generation \cite{snodgrass2013generating} it make sense to apply it on this new task of level inpainting. 
We use the original formulation of the Markov Chain from \cite{snodgrass2013generating} trained on the same levels that make up the training set for our two models.
During testing, we query the model to fill in only the masked section of the input.

\section{Results}

We include the average and standard deviation values of our three metrics in Table \ref{tab:t1} for the convolutional autoencoder (AE), the U-net (UNet), and the Markov approach (Markov). 
Each row represents a particular model's results for one of our three different metrics.
For each column we have each test level, and then a final column giving the average over these values.

In the first three rows, we can see the results of our different implementations evaluated using the Tile-by-Tile (TbyT) metric. Of these three implementations we can see that the convolutional autoencoder and the U-net achieve similar results compared to the Markov implementation. 
In addition, the convolutional autoencoder consistently outperforms the U-net.
We anticipate this was due to the convolutional autoencoder being a simpler model than the U-net, and so it did a slightly better job generalizing over the same data.

\begin{figure*}[tbh]
    \centering
    \includegraphics[width=\textwidth]{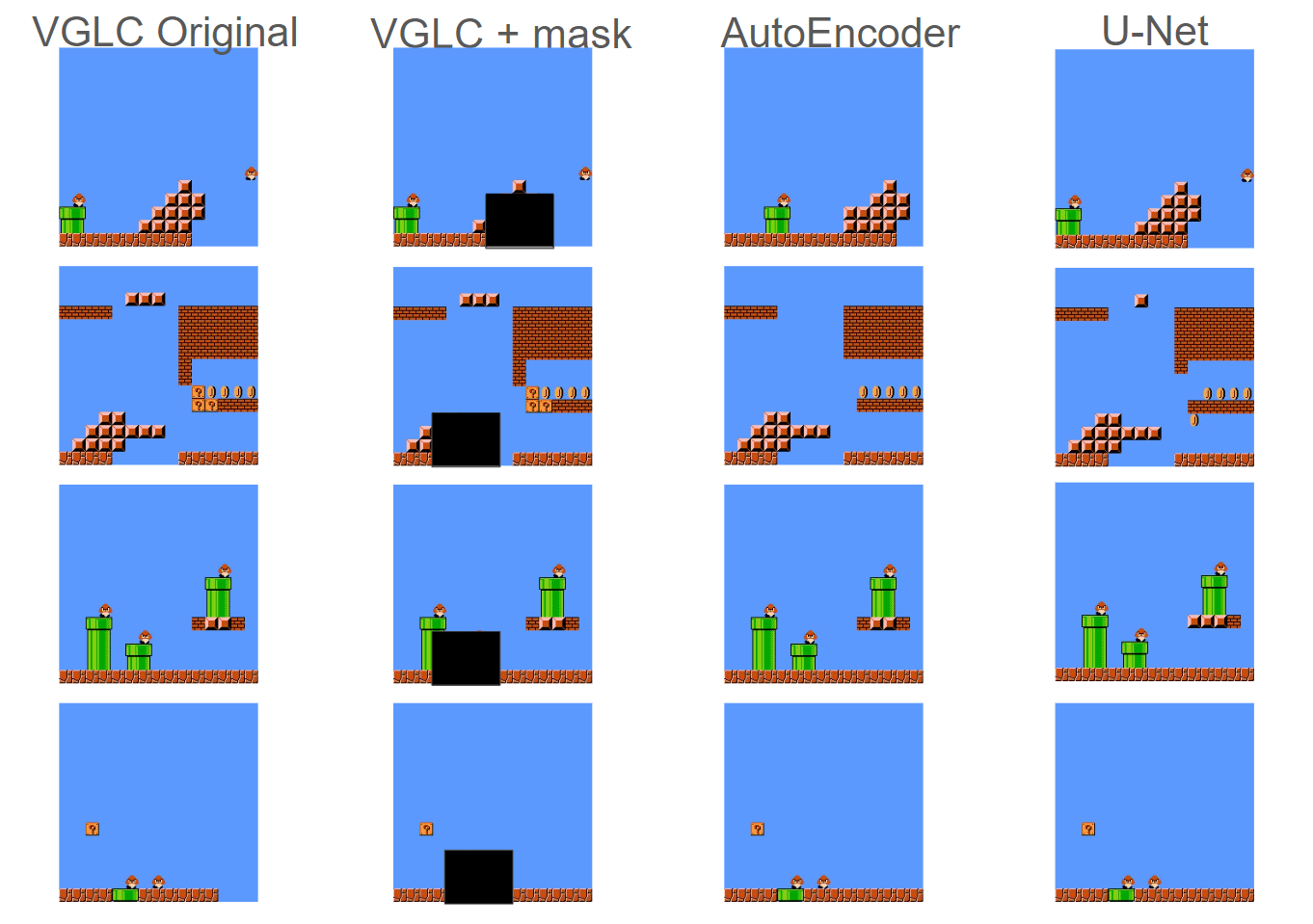}  
    \caption{In this image, we include original level sections, one of the masks added to the level section, and the output of the autoencoder and the U-net. These examples were selected randomly from the test dataset.}
    \label{fig:complete}
\end{figure*}

In the middle three rows of Table \ref{tab:t1} we can see the results using the No Sky metric (NoSky), in which we exclude the sky tiles. 
All models did slightly worse in this case, indicating an over-reliance on predicting sky tiles.
In these results we can see that the autoencoder again had a slight advantage in the reconstruction accuracy. 

For our last three rows, we used the Structures (Struct.) metric. 
With this metric we check the tile accuracy only in terms of pipes and stairs. 
As we can see the U-net ended up with a small advantage compared to the autoencoder. 
We anticipate this is because the U-net is better at reproducing less common structure \cite{ronneberger2015u}.
In comparison, the Markov chain had the worst results, unable to produce any pipes or stairs for most levels. It consistently performed worse compared to the other models, suggesting that even techniques that work well for level generation may not be suitable for content augmentation or specifically level inpainting.

In Figure \ref{fig:complete} we include some of the output generated by the autoencoder and the U-net in comparison to the original VGLC dataset.
All of these examples were taken from the test set.
The performance is close, which matches the quantitative results in the table. 
However, the U-net seems to have done a slightly better job at capturing the location of the more unique tiles, such as the stairs in the first two examples and the question mark block in the third example.

\section{Case Study}
\begin{figure*}[tbh]
    \centering
    \includegraphics[width=\textwidth]{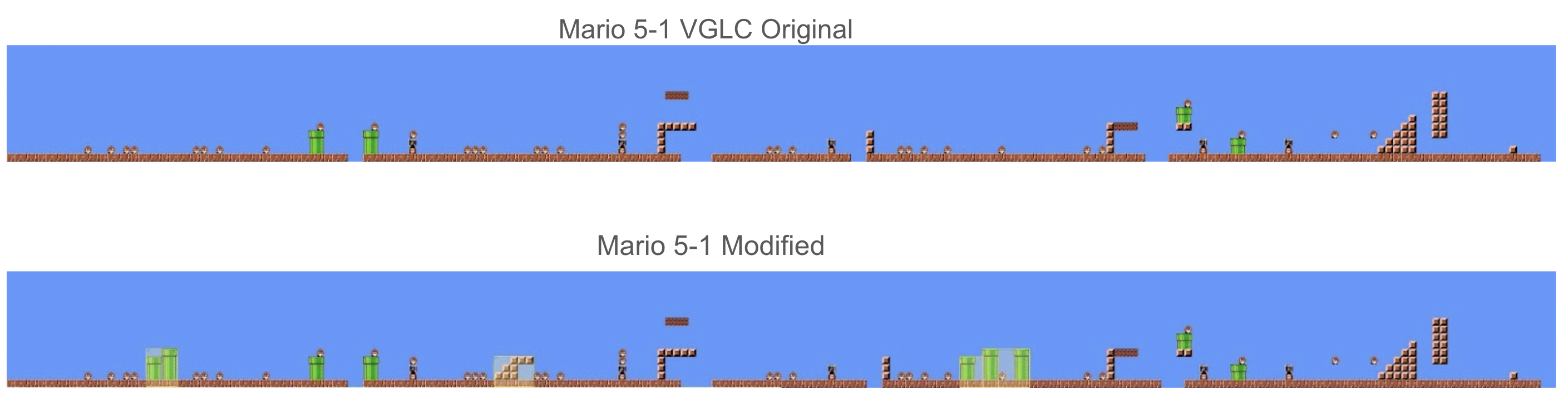}    
    \caption{Example based on \textit{Super Mario Bros.} 5-1, at the top we include the original level from VGLC. At the bottom, we present a new version produced with Content Augmentation in the yellow areas, applying only new structures with level inpainting.}
    \label{fig:extraM}
\end{figure*}

In the last section we demonstrated that a convolutional autoencoder outperforms a U-net for almost all the metrics and both models consistently outperform a baseline Markov chain. 
We take this to indicate that both of these approaches can successfully be applied for level inpainting.
However, we did not show it applied to our high-level task of Content Augmentation (CA) due to the lack of an ability to quantitatively evaluate any output. 

As an illustrative example of this initial approach, we demonstrate what one version of augmenting a level might look like in Figure \ref{fig:extraM}. In this figure we present the original map of the classic \textit{Super Mario Bros.} 5-1 and an updated version where the content has been augmented.
We did this manually by selecting areas with low structural content (just sky and ground) and applying a mask size where new structures were placed by using the U-net model. 
One could imagine this as the output of a co-creative level inpainting tool or a ``version 2.0'' of the level output by a more autonomous tool.

\section{Limitations}

We acknowledge that our approach taken in this paper has a number of limitations. 
For example, the VGLC dataset does not contain all the underground levels or the castle levels from the original games. 
Consequently, it is impossible to reconstruct these types of structures for our model.
Other limitations to consider are the position and shape of the mask. Given that there are levels where the structures are not only along the bottom of the level, it is not possible to evaluate the reconstruction of some structures. One way to address this issue could be to have the mask follow the player's path. With this, we might end up with a model that can better reproduce structures important to players, similar to the finding from Summerville and Mateas \cite{summerville2016super}. 
Despite these limitations, we consider that our results are still valid for the following reasons. First, the mask that we implemented covered the majority of the structures at the bottom of the levels. Second, most of the levels had content along the bottom. Finally, our models showed a reasonable reconstruction of the level structures.

For this initial exploration of level inpainting we adapted two models previously used for image inpainting. 
While both models outperformed a PCGML baseline, their quantitative performance and qualitative performance (as demonstrated in Figure \ref{fig:extraM} could both be improved. 
Though we feel this is appropriate for an initial exploration, there are definitely avenues to address this limitation through future work.

\section{Future Work}

For future work, we plan to focus on different aspects of Content Augmentation (CA). 
There are a number of different options, with one being to improve the quality of level inpainting. 
One possible way to improve level inpainting would be to continue trying to adapt different systems from image inpainting. 
Image inpainting has a significant amount of prior work focused on finding an optimal mask shape. 
Similarly, we could explore the choice of mask shape in more detail. 
As an alternative to focusing on adapting image inpainting models to level inpainting, we could adapt additional PCGML models for level generation to this task. 
For example, WaveFunctionCollapse (WFC) \cite{Sandhu2019}, which could be adapted to level inpainting given the way it iteratively generates output content. 
The key would be to treat the masked out content as yet-ungenerated content, and the areas around the mask as content previously generated by WFC. 
Though it may be necessary to handle cases where the existing contradicts the constraints WFC has extracted from existing levels.

Outside of improving level inpainting we hope to explore other aspects of CA.
In particular, aspects related to game mechanics.
The concept would be to use CA to generate additional abilities, items, or enemies for an existing game. 
The goal would be to get output similar to what one might find in human-authored DLC, mods, or updates in a live service game.
Other possible options might be to focus on particular level design applications in CA. For example, merging different levels via level inpainting, or identifying the best locations in a map to add new content while keeping a coherent game design.
In the future, we hope that CA can lead to systems to automatically extend existing games, producing DLC, mods, or entirely new kinds of game designs.

\section{Conclusions}

We presented a new PCG problem, Content Augmentation (CA), and how it relates to classical PCG and PCGML. 
We introduced a subproblem of CA: level inpainting, where the main idea is to reconstruct missing information in a level. Our experiments demonstrate that image inpainting architectures can be adapted for level inpainting. 
Despite this, these adaptations seem to require more modifications to improve their performance for CA. 
Given these initial positive results, we plan to continue this line of research towards generation of novel CA in the future.

\section*{Acknowledgements}

This work was funded by the Canada CIFAR AI Chairs Program, Alberta Machine Intelligence Institute, and the Natural Sciences and Engineering Research Council of Canada (NSERC).

\bibliography{aaai23}

\end{document}